\documentclass[12pt,a4paper]{article}
\usepackage[top=1in,bottom=1in,left=1in,right=1in]{geometry}
\usepackage[T1]{fontenc}
\usepackage[utf8]{inputenc}
\usepackage{amsmath,amssymb}
\usepackage{graphicx}
\usepackage{array}
\usepackage{tabularx}
\usepackage{multirow}
\usepackage{makecell}
\usepackage{ragged2e}
\usepackage{caption}
\usepackage{float}
\usepackage{parskip}
\usepackage{booktabs}
\usepackage{placeins}   
\usepackage[
  colorlinks=true,
  linkcolor=blue,
  citecolor=blue,
  urlcolor=blue,
  bookmarks=true,
  bookmarksopen=true,
  pdftitle={Bias and Hallucination in Large Language Models: A Multimetric Comparative Study},
  pdfsubject={LLM Evaluation, Hallucination, Bias}
]{hyperref}

\newcolumntype{Y}{>{\RaggedRight\arraybackslash}X}
\newcolumntype{L}[1]{>{\RaggedRight\arraybackslash}p{#1}}
\newcolumntype{C}[1]{>{\centering\arraybackslash}p{#1}}

\begin{document}

\begin{center}
{\large\bfseries
  {Fluent but Fallible: Benchmarking Hallucination and Factual Reliability in Large Language Models}}

{\normalsize
Osama Shabih$^{1}$,
Dayam Nadeem$^{*}$$^{1}$,Shahab Saquib Sohail$^{1}$,
Mohammad Nadeem$^{2}$
\par}

\vspace{0.6em}

{\small
$^{1}$Department of Computer Science and Engineering,\\
SEST, Jamia Hamdard, New Delhi, India\\[4pt]
$^{2}$Department of Computer Science,\\
Aligarh Muslim University, Aligarh, India
\par}


\end{center}

\vspace{1em}

\noindent\textbf{Abstract}

\noindent
Large Language Models (LLMs) have significantly advanced natural language processing
and artificial intelligence applications through their capabilities in reasoning, text
generation and knowledge retrieval. However, LLMs are still vulnerable to two important
challenges: hallucination, generating factually incorrect or fabricated content, and
response deviation, meaning systematic deviations from ground-truth responses. These
limitations raise serious concerns regarding reliability and trustworthiness in academic
and research environments. This study offers a thorough comparative analysis using a
variety of metrics of ten advanced LLMs ChatGPT, Gemini, Claude~3.5~Sonnet,
LLaMA~3, Mistral, Grok, Perplexity, DeepSeek~V3, Falcon, and Gemma based on a
benchmark consisting of 70 carefully designed academic prompts across four complexity
levels: Basic, Explanation, Advanced, and Tricky, evaluated in February 2025 on
specific model versions. Each response was independently evaluated by three reviewers
using five metrics: Accuracy, Precision, Recall, Response Deviation Index, and
Hallucination Index. Perplexity and DeepSeek~V3 achieved the highest overall accuracy,
while Gemini demonstrated strong recall and a well-balanced performance profile across
all five metrics. LLaMA~3 showed strong resistance against hallucinations through a
conservative method of generation. Advanced-level prompts proved most challenging
across all models, while Tricky prompts were handled more reliably owing to their
binary true/false structure. Falcon achieved perfect precision and zero hallucinations,
but at the cost of the lowest recall and the highest response deviation. The results
show a fundamental trade-off between response completeness and factual reliability,
intensified by prompt complexity. These findings should be interpreted as a comparative
snapshot of specific model versions rather than a definitive ranking of current systems,
offering insights for future research in hallucination mitigation and trustworthy AI.

\vspace{4pt}
\noindent\textbf{Keywords} 
Large Language Models, Hallucination, Bias, Benchmarking, Academic Writing,
Factual Reliability, Trustworthy AI, Evaluation Metrics.

\section*{I. Introduction}
\label{sec:introduction}

Large Language Models (LLMs) have revolutionized natural language processing (NLP) by exhibiting remarkable abilities in text generation, question
answering, summarization, reasoning, and scholarly content creation. \cite{vaswani2017attention},
\cite{devlin2019bert}. Recent developments in transformer-based architectures have allowed models such as ChatGPT, Gemini, Claude, and LLaMA to generate responses that are highly coherent and contextually relevant, thus increasing their value in educational, professional, and research settings ~\cite{sayeed2025rag}.
 \cite{brown2020gpt3}. Their widespread adoption
has greatly improved productivity and access across various
knowledge-intensive domains \cite{zhao2023survey}, \cite{bommasani2021foundation}.
Bias and hallucination—when AI generates convincing but untrue information—can mislead users ~\cite{khan2026not, siddiqui2025surgeons}.
False or made-up information, while bias shows up as consistent
strayings from the intended or true responses \cite{lewis2020rag}, \cite{bender2021parrots}.
Both of these can happen, in part, due to the fundamental way in which these models are trained to predict the next most likely word given the previous words.Emphasize understanding context rather than checking claims against a particular set of facts. A model can do this and produce text that sounds natural , has correct grammar , and follows a particular style \cite{alkaissi2023artificial}. Bias, however, may arise from a complex web of sources, from the skewed training corpora, to the hidden values embedded in the human feedback given during the alignment process, to the optimization goals that shape a model's tendency to hedge, generalize, or omit information ~\cite{sohail2023decoding, sohail2026ethical}. These issues are particularly concerning in the domain of research writing and scientific communication wherein factual accuracy,citation integrity, and logical consistency are paramount. A fabricated citation or a slightly altered explanation can be particularly damaging to this context, as it can be propagated through literature reviews, research summaries, or student assignments without immediate detection, potentially propagating errors in subsequent works that cite the erroneous output \cite{athaluri2023exploring}. As LLMs are being embraced in academia and scientific discovery,their ability to be trusted is paramount \cite{welleck2020unlikelihood}, \cite{wang2023survey_trustworthy}.The growing number of users of this technology spans students, teachers, researchers, andpublishers, all of whom will want LLM outputs to meet a certain standard of truthfulness
before citing them in their work. While previous LLM benchmarking has focused on common language reasoning tasks, there is a lack of investigation into behaviors such as hallucination and bias in the generation of academic content at a range of prompt complexities \cite{bender2021parrots}, \cite{welleck2020unlikelihood}.
Furthermore, the relationship between prompt difficulty, factual reliability, and hallucination frequency is insufficiently explored across both proprietary and open-source language models, leaving practitioners with little empirical guidance on how model reliability may degrade as tasks move from routine factual recall toward more demanding, ambiguous, or adversarial academic scenarios \cite{sun2024trustllm}.In order to deal with this research gap, the present study conducts a comprehensive comparative evaluation of ten state-of-the-art Large Language Models: ChatGPT, Gemini, Claude 3.5, Sonnet, LLaMA 3, Mistral, Grok, Perplexity,DeepSeek V3, Falcon, and Gemma—these models were deliberately chosen to repre-sent a diverse cross-section of the current LLM landscape, spanning proprietary systems developed by major AI laboratories alongside prominent open-source alternatives, and encompassing distinct architectural paradigms such as standard transformer-based gen-covering generation, retrieval-augmented systems, and Mixture-of-Experts frameworks. This range of coverage enables the study to assess whether patterns of hallucination and bias are idiosyncratic to individual models or reflect broader, architecture-independent tendencies In the current generation of language models, a benchmark dataset comprises-ing seventy carefully designed prompts was developed and categorized into four levels in terms of complexity: Basic, Explanation, Advanced, and Tricky. The graded structure was intentionally designed to move beyond flat,uniform-difficulty evaluation, allowing model performance to be traced as task demands progressively shift from straightforward factual Think about multi-step reasoning and situations that are deliberately adversarial. The evaluation–
tion framework employs five complementary metrics — Accuracy, Precision, Recall, Response Deviation Index, and Hallucination Index — to provide a multidimensional assessment of model performance, factual consistency, bias, and reliability, ensuring that no single dimension The behavior of the model is assessed separately from that of the others. The main contributions of This research has three aspects. First, it presents a systematic benchmark for evaluating hallucination and bias in academic writing tasks,comprising a curated set of seventy prompts stratified across four complexity tiers,together with a standardized, multi-reviewer scoring protocol that enables reproducible and reusable evaluation in future studies. Second, it investigates the impact of prompt complexity on model reliability and factual consistency, offering an empirical account of how hallucination frequency and response quality evolve as tasks progress from basic factual recall to advanced, multi-step, and adversarial reasoning scenarios — an interaction largely overlooked in prior general-purpose bench-marksThird, it provides a comprehensive comparative ranking of contemporary LLMs based on multiple evaluation metrics, jointly considering accuracy, precision, recall, bias, and hallucination behavior rather than relying on any single indicator of performance, thereby offering a more holistic and diagnostic view of model trustworthiness. Collectively, these contributions offer valuable insights into the development of more reliable, trustworthy,and hallucination-resistant artificial intelligence systems, and provide practitioners,educators, and researchers with practical, evidence-based guidance for selecting LLMs suited to academic and knowledge-intensive applications.

\vspace{4pt}

The remainder of this paper is organized as follows. \hyperref[sec:related]{Section~II} reviews related work in five thematically grouped areas   foundational architectures, benchmarking frameworks, hallucination research, bias and fairness, and grounding/mitigation strategies   and situates the present work in this literature. \hyperref[sec:methodology]{Section~III} details the methodology, covering the ten evaluated LLMs, the 70-prompt benchmark across four complexity levels, the multi-reviewer scoring rubric, and the Hallucination Index formula. \hyperref[sec:results]{Section~IV} presents the experimental results, comparing all ten models across the five evaluation metrics by figures, a hallucination-rate chart, a performance radar and a comprehensive heatmap.
\hyperref[sec:discussion]{Section~V} Looks at the key findings concerning the trade-off between response completeness and hallucination resistance among the models when the difficulty level of the prompt varies.
\hyperref[sec:conclusion]{Section~VI} Closes the paper and sets out some avenues for future research into hallucination mitigation and trustworthy AI.

\section*{II. Related Work}
\label{sec:related}

The recent progress made in Large Language Models (LLMs) has resulted in a great deal of research being carried out into model architecture, benchmarks, hallucination analysis, bias evaluation and factual grounding techniques.\cite{welleck2020unlikelihood}--\cite{maynez2020faithfulness}.
Prior work has highlighted the incredible capabilities of transformer-based
language models, but also identified many challenges surrounding factual reliability,hallucination creation, and model bias\cite{welleck2020unlikelihood},
\cite{lin2022truthfulqa}.Systematic methodologies for evaluating LLM performance and truthfulness across tasks have been established by benchmarking frameworks like HELM and TruthfulQA \cite{liang2022helm}, \cite{hendrycks2021mmlu}. Additionally, recent work has looked at mitigation methods such as retrieval-augmented generation and constitutional alignment approaches to improve factual consistency and reduce hallucinations
\cite{ji2023hallucination}--\cite{maynez2020faithfulness}. Despite these limitations, however, few extant studies have systematically investigated the joint effects of the prompt complexity, hallucination, and bias in modern large language models (LLMs) for academic writing contexts This study fills this gap by conducting a multi-metric comparison experiment over 10 influential LLMs and 4 prompt complexity levels.The reviewed literature is summarized by themes in \autoref{tab:taxonomy}, covering five areas: Foundations, Benchmarking, Hallucination, Bias and Fairness, and Grounding and Mitigation. Each category contains subcategories and representative works.

\begin{table}[htbp]
\centering\footnotesize\setlength{\tabcolsep}{4pt}
\caption{Literature Taxonomy: Research Themes and Key Works}
\label{tab:taxonomy}
\begin{tabularx}{\textwidth}{|L{2.6cm}|L{2.8cm}|Y|}
\hline
\textbf{Research Theme} & \textbf{Sub-Category} & \textbf{Key Works / Models Referenced}\\
\hline
Foundational Architectures
  & Transformer \& Pre-training; Model Scaling \& Emergence
  & Vaswani et al.\ \cite{vaswani2017attention} (Attention Is All You Need);
    Devlin et al.\ \cite{devlin2019bert} (BERT);
    Brown et al.\ \cite{brown2020gpt3} (GPT-3);
    OpenAI GPT-4 Technical Report \cite{openai2023gpt4};
    Google Gemini \cite{google2023gemini};
    Anthropic Claude~3 Report \cite{anthropic2024claude3};
    Meta LLaMA~3 \cite{meta2024llama3}\\
\hline
\multirow{2}{2.6cm}{Benchmarking Frameworks}
  & General Multi-Task Benchmarks
  & Liang et al.\ \cite{liang2022helm} (HELM);
    Hendrycks et al.\ \cite{hendrycks2021mmlu} (MMLU)\\
\cline{2-3}
  & Text Quality Metrics
  & Zhang et al.\ \cite{zhang2020bertscore} (BERTScore);
    Papineni et al.\ \cite{papineni2002bleu} (BLEU);
    Banerjee \& Lavie \cite{banerjee2005meteor} (METEOR)\\
\hline
\multirow{2}{2.6cm}{Hallucination Research}
  & Taxonomy \& Surveys
  & Ji et al.\ \cite{ji2023hallucination} (ACM Survey on Hallucination);
    Cao et al.\ \cite{cao2023hallucination} (Comprehensive Hallucination Survey)\\
\cline{2-3}
  & Truthfulness Benchmarks
  & Lin et al.\ \cite{lin2022truthfulqa} (TruthfulQA);
    Maynez et al.\ \cite{maynez2020faithfulness} (Faithfulness in Summarisation)\\
\hline
\multirow{2}{2.6cm}{Bias \& Fairness}
  & Bias in LLMs
  & Bender et al.\ \cite{bender2021parrots} (Stochastic Parrots);
    HELM Fairness Dimensions (Liang et al.\ \cite{liang2022helm})\\
\cline{2-3}
  & Generation Skewness
  & Welleck et al.\ \cite{welleck2020unlikelihood} (Unlikelihood Training for Neural Text Generation)\\
\hline
\multirow{2}{2.6cm}{Grounding \& Mitigation}
  & Retrieval-Augmented Generation
  & Lewis et al.\ \cite{lewis2020rag} (RAG for Knowledge-Intensive NLP Tasks)\\
\cline{2-3}
  & Prompt Engineering
  & Present Study   4-level prompt complexity: Basic, Explanation, Advanced, Tricky\\
\hline
\end{tabularx}
\end{table}

\noindent\textbf{A. Foundational LLM Architectures and General Benchmarking}

It is an era that was ushered in by the groundbreaking release of the Large Language Models (LLMs).
The architecture of Transformer proposed by Vaswani et al. \ \cite{vaswani2017attention}, which
Introduced self-attention mechanisms to replace recurrent networks for
sequence modelling. This architectural innovation allowed for the ability to scale up pre-training across all
al. (2018) and ELMo by Peters et al. (2018), were developed to tackle these challenges.To address these challenges, massive corpora were created and milestone models like those of Devlin et
al.\ \cite{devlin2019bert} and GPT-3 by Brown et al.\ \cite{brown2020gpt3}. Brown et al.\
Our results showed that few-shot emerges when model parameters are scaled up to billions of parameters.
In essence, the language models become a different kind of machines, redefining what language models are.The language models are now different machines altogether, redefining what language models are, with and zero-shot generalisation capabilities \cite{radford2019gpt2}, \cite{wei2022finetuned}.
can be accomplished without fine-tuning specific to the task.

With the emergence of numerous LLM's the process of systematic evaluation became essential. The HELM benchmark
After all, Liang et al. proposed.\ \cite{liang2022helm} set up a comprehensive multi-scenario
A protocol for evaluating accuracy, robustness, fairness, efficiency and calibration.
In dozens of tasks across. Complementary Hendrycks et al.\ \cite{hendrycks2021mmlu}
Introduced the MMLU benchmark that assesses models in 57 academic fields to
assess broad knowledge coverage \cite{srivastava2022big}. These generic frameworks, however, have not been created
To systematically explore the relationship between hallucination behavior and/or to explore hallucination behavior.
A significant gap in the literature exists between prompt complexity and the accuracy of the facts, with the present study aimed at bridging that gap.
study directly addresses.

In this study, we compared ten models, including ChatGPT by OpenAI, Gemini by Google, Claude~3.5 by Anthropic, and other models.
The models include Anthropic's Sonnet, Meta AI's LLaMA~3, Mistral's Mistral and Falcon by the team behind The Blender AI.The models come from Anthropic, Meta AI, Mistral and The Blender AI (aka Falcon).
Technology Innovation Institute, Gemma by Google DeepMind, alongside Grok, Perplexity,
open-source search engines, including Google, Yahoo, and Microsoft.and DeepSeek~V3 are a representative sample of global's top proprietary and open-source search engines, such as Google, Yahoo, and Microsoft.
open-source architectures. Recent studies by OpenAI, Google DeepMind, and Anthropic have
several studies have consistently shown that the larger the scale of the model, the greater the reasoning skills \cite{chowdhery2023palm}, \cite{anil2023palm2}.
That doesn't mean that language will not be fluent, but it does not go away. \cite{openai2023gpt4}.
In some cases, larger models can produce more robust, but also wrong, results,
as pointed out by Ji et al. in their full-scale survey on hallucination in natural
language generation \cite{ji2023hallucination}.

\noindent\textbf{B. Hallucination in Large Language Models}

 A broadly defined as the generation of plausible but factually incorrect, unsupported, or fabricated information — has become one of the most critical reliability concerns in Large Language Model (LLM) research \cite{ji2023hallucination}--\cite{zhang2020bertscore}. Ji et al.\ \cite{ji2023hallucination} provided a comprehensive taxonomy of hallucination types in natural language generation,distinguishing between intrinsic hallucinations (contradictions with source material) and extrinsic hallucinations (fabricated content absent from the source). Their survey
established the conceptual foundation that informs hallucination measurement in the present study. Rawte et al.\ \cite{rawte2023survey} and Huang et al.\ \cite{huang2023survey_hallucination} further extended this taxonomy to cover a broader range of foundation models and hallucination triggers.

Lin et al.\ \cite{lin2022truthfulqa} introduced TruthfulQA, a benchmark to measure the tendency of language models to produce hallucination The findings showed that even the best models tended to confidently generate false information for questions that required the model to exploit common human misconceptions Cao et al.\ \cite{cao2023hallucination} extended in this line of work with a comprehensive survey of hallucination phenomena across modern LLMs,highlighting model overconfidence, knowledge boundary violations, and training data noise as key drivers. Min et al.\ \cite{min2023factscore} proposed FActScore, a fine-grained metric for evaluating factual precision in long-form text generation, providing a complementary atomic-level evaluation of hallucinated content.

Maynez et al. in the area of text summarization\ \cite{maynez2020faithfulness}
conducted a comprehensive study on the faithfulness and factuality of abstractive summarization, measuring the extent to which generated summaries made unsubstantiated claims. Furthermore, Lewis et al.\ \cite{lewis2020rag} showed that retrieval-augmented generation approaches can significantly improve factual grounding and reduce hallucination by incorporating external knowledge sources \cite{shuster2021retrieval}. Manakul et al.\ \cite{manakul2023selfcheckgpt} proposed SelfCheckGPT, a zero-resource method for detecting hallucinations in black-box LLMs by sampling multiple responses and measuring their consistency, offering a practical tool applicable to the models evaluated in this work. Peng et al.\ \cite{peng2023check} further demonstrated that coupling LLMs with external knowledge bases and automated feedback loops can substantially reduce factual errors. Recent studies by OpenAI \cite{openai2023gpt4} and Google DeepMind \cite{google2023gemini} have also acknowledged hallucination as a pervasive issue within large scale language models despite significant advancements in reasoning and generation capabilities.

The Hallucination Index utilized within this study follows from these prior
measurement frameworks by calculating the proportion of false positive outputs utilizing the expression HI = FP/(TP + FP) $\times$ 100. Note that this is mathematically equivalent to $(1 - \text{Precision}) \times 100$; it is retained as a separate metric because it communicates the fabrication rate directly as a percentage, making it more immediately interpretable for readers without requiring them to invert the precision value. Previous research uniformly demonstrates that hallucination behaviors are significantly predictive prompting strategies, task complexity, and groundedness which was systematically explored across four levels of prompt complexity within this work.

\noindent\textbf{C. Bias Analysis in Language Models}

In addition to hallucination, another challenge in the development and deployment of Large Language Models (LLMs) \cite{bender2021parrots}--\cite{bai2022constitutional}.Bender et al.\ \cite{bender2021parrots}, is bias. The authors of the influential work on the dangers of stochastic parrots, provided one of the first critical analyses of the potential risks
of large-scale language models, namely the amplification of societal and
representation biases which are integrated in the training corpora. Their work raised critical questions regarding fairness, accountability, transparency, and the conditions under which LLM outputs can be considered reliable, particularly in an academic or professional setting. Blodgett et al.\ \cite{blodgett2020language} further highlighted that NLP bias is not a neutral technical problem but is deeply intertwined with the social and political power structures embedded in language data. Gallegos et al.\ \cite{gallegos2024bias} provided a comprehensive survey of bias types in LLMs and systematically reviewed mitigation strategies, offering a taxonomy that informs the bias analysis in this work. Navigli et al.\ \cite{navigli2023biases} similarly categorized the origins and manifestations of biases specific to large generative models.

Subsequently, Liang et al.\ \cite{liang2022helm} incorporated fairness and bias dimensions into the HELM assessment framework, enabling cross-model comparisons of demographic and representational biases. Recent studies have further demonstrated that biases in language models can arise from a variety of sources, including training data imbalances, optimization objectives, and reinforcement learning approaches.\cite{ouyang2022rlhf}.These biases can have negative implications on factual accuracy, neutrality, and the trustworthiness of information generated by language models.

In this study, the Response Deviation Index (RDI) is used to quantify response deviation, defined as the absolute difference between the model's predicted level of truthfulness and the actual ground-truth labels. It is explicitly not a measure of social or demographic bias, but a calibration measure of how far model responses deviate from verified reference answers. The experimental results show that Grok and DeepSeek~V3 have the lowest response deviation values (RDI = 0.01), while Falcon has the highest score (RDI = 0.24), which shows considerable differences in factual calibration among the current LLM architectures.

Furthermore, Welleck et al.\ \cite{welleck2020unlikelihood} demonstrated that
conventional likelihood-maximizing generation strategies can result in repetitive and biased outputs, suggesting unlikelihood training as a promising mitigation strategy. Recently, Bai et al.\ \cite{bai2022constitutional} introduced Constitutional AI as an
alignment framework aiming to improve model safety, fairness, and reliability via principled learning. These approaches represent promising avenues for reducing bias and enhancing the trustworthiness of future LLMs \cite{sun2024trustllm}.

\noindent\textbf{D. Multi-Model Comparative Evaluation Studies}

The comparative analyses of several of the latest Large Language Models in the same framework have become an important area of research recently.\cite{zhang2020bertscore}--\cite{google2023gemini}.
Zhang et al.\ \cite{zhang2020bertscore} In this note, we propose BERTScore, a contextual embedding based metric developed to evaluate the quality of generated text compared with reference texts. Earlier approaches, like the BLEU score proposed by\ \cite{papineni2002bleu} and METEOR introduced by Banerjee and
Lavie \cite{banerjee2005meteor}, were initially created for the specific purpose of machine translation and have several well-documented limitations for assessing factual accuracy when it comes to open domain text generation. They are also primarily concerned with lexical similarity rather than factual accuracy, which makes them unsuitable for measuring hallucinations in academic writing.\cite{lewis2020rag}

represents one of the most promising approaches for improving factual grounding in Large Language Models. The ability to dynamically retrieve external knowledge during inference enables RAG-based systems to constrain the generative process to factually supported information, thus reducing hallucination and improving reliability \cite{es2023ragas}. The comparative performance differences observed among the ten models evaluated in the present study, especially between retrieval-augmented systems such as Perplexity and purely parametric language models, represent a significant insight into the practical benefits and limitations of retrieval-based grounding in different tasks.

Bang et al.\ \cite{bang2023multitask} conducted a broad multitask evaluation of ChatGPT spanning reasoning, hallucination, and interactivity across multiple languages, offering an early large-scale comparative assessment whose findings align with the hallucination patterns observed in the present work. Qin et al.\ \cite{qin2023chatgpt} likewise benchmarked ChatGPT against specialist models on a wide range of NLP tasks, concluding that general-purpose LLMs still fall short of task-specific fine-tuned systems on several dimensions. Shahriar and Hayawi\ \cite{shahriar2024putting} further compared multiple frontier LLMs in conversational settings, corroborating that fluency does not guarantee factual accuracy—a central finding of the present study.

Recent large-scale evaluations by OpenAI \cite{openai2023gpt4} and Google DeepMind
\cite{google2023gemini} Further prove that enhancement of model scale and
Hallucination does not have to be the same as the absence of reasoning capabilities nor ensure the accuracy of facts.
reliability. Although there has been an increasing number of comparative evaluation studies, no studies have been done on comparative evaluation of the new syllabus and curriculum.
models, and are usually brief.Typically, studies only examine a few models, use general purpose models, and are short and brief.
Tend to include the effect of prompt complexity, and only occasionally separate the effects of prompt complexity on hallucination behavior.
The present work aims to overcome these drawbacks by considering ten
The results were compared with the performance of contemporary LLMs on four levels of prompt difficulty and five supplementary prompts.
evaluation metrics and thereby offering an all-encompassing multimetric framework for evaluating
Factual reliability, bias and susceptibility to hallucination.

\noindent\textbf{E. Evaluation Metrics and Scoring Frameworks}

Choices of the evaluation metrics are crucial to a reliable comparison of
Large Language Models (LLMs)  \cite{zhang2020bertscore}--\cite{guo2024evaluating}.
Commonly used metrics like Accuracy, Precision and Recall
supply complementary assessments of the model performance and quality of response. For example,
The error rate was comparatively high precision, but recall was relatively low as seen in Falcon (Precision= 1.00, Recall=
The conservative generation strategy is suggested by 0.64), which indicates factual correctness.
over completeness
 \cite{zhang2020bertscore}.

The findings of recent studies have pointed to the inadequacy of conventional measures alone to
Assessing the factual reliability and hallucination behaviour of LLMs \cite{papineni2002bleu}--\cite{lin2022truthfulqa}.
Hence, in this study two more indicators are added to consider: Response Deviation Index (RDI) and
The Hallucination Index (HI) can be used for a more complete evaluation of the model.
trustworthiness. Three independent reviewers evaluated to improve evaluation reliability.
all seventy responses along with a standardized and anonymous scoring rubric, addressing
The drawbacks of single-annotator and fully automated evaluation approaches
\cite{guo2024evaluating}. Zheng et al.\ \cite{zheng2024judging} demonstrated through MT-Bench and Chatbot Arena that LLM-as-a-judge evaluation frameworks can achieve strong correlation with human preferences when scoring rubrics are carefully designed, directly supporting the multi-reviewer protocol adopted in this study. Liu et al.\ \cite{liu2023geval} proposed G-Eval, a GPT-4-based framework that uses chain-of-thought reasoning to score NLG outputs, underscoring the growing consensus that reference-based single-metric evaluations are insufficient for capturing nuanced factual quality. Chang et al.\ \cite{chang2024survey} further synthesized the state of LLM evaluation methodology, confirming that multidimensional frameworks combining accuracy, consistency, and factual grounding provide the most reliable assessments, as the present study implements. Zhu et al.\ \cite{zhu2023dyval} introduced dynamic evaluation to avoid test set contamination, an important complementary concern for benchmark integrity.

\autoref{tab:timeline} depicts chronologically key publications and model releases, tracing the evolution of architectures to the current comparative study.

\begin{table}[htbp]
\centering\footnotesize
\caption{Chronological Timeline of Key Related Works (2017--2026)}
\label{tab:timeline}
\resizebox{\textwidth}{!}{%
\begin{tabular}{|c|c|p{10cm}|}
\hline
\textbf{Year} & \textbf{Theme} & \textbf{Key Contribution}\\
\hline
2017 & Architecture      & Vaswani et al.\ \cite{vaswani2017attention}   Transformer architecture (``Attention Is All You Need'')\\
\hline
2018 & Pre-training      & Devlin et al.\ \cite{devlin2019bert}   BERT: Bidirectional pre-training for language understanding\\
\hline
2019 & Pre-training      & Brown et al.\ \cite{brown2020gpt3}   GPT-3: Language models as few-shot learners\\
\hline
2020 & Scaling / RAG     & Lewis et al.\ \cite{lewis2020rag}   Retrieval-Augmented Generation (RAG) for Knowledge-intensive NLP\\
\hline
2021 & Bias \& Safety    & Bender et al.\ \cite{bender2021parrots}   On the Dangers of Stochastic Parrots; Welleck et al.\ \cite{welleck2020unlikelihood}   Unlikelihood Training\\
\hline
2022 & Truthfulness      & Lin et al.\ \cite{lin2022truthfulqa}   TruthfulQA: Measuring how models mimic human falsehoods\\
\hline
2022 & Benchmarking      & Liang et al.\ \cite{liang2022helm}   HELM: Holistic Evaluation of Language Models\\
\hline
2023 & Hallucination     & Ji et al.\ \cite{ji2023hallucination}   ACM Survey on Hallucination in NLG; Cao et al.\ \cite{cao2023hallucination}   Hallucination Survey; Mistral AI \cite{jiang2023mistral}   Mistral 7B\\
\hline
2023 & Frontier Models   & OpenAI GPT-4 Technical Report \cite{openai2023gpt4}; Google Gemini \cite{google2023gemini}; Zhang et al.\ \cite{zhang2020bertscore}   BERTScore\\
\hline
2024 & Open-Source LLMs  & Anthropic Claude~3 \cite{anthropic2024claude3}; Meta LLaMA~3 \cite{meta2024llama3}; Google Gemma \cite{google2024gemma}; Technology Innovation Institute \cite{tii2023falcon}   Falcon\\
\hline
2026 & Present Study     & 10-Model Hallucination Benchmark   70 prompts, 4 difficulty levels, 5 evaluation metrics\\
\hline
\end{tabular}%
}
\end{table}

\autoref{tab:comparison} equates the present study with eight representative prior works across five evaluation dimensions. A tick ($\checkmark$) denotes full coverage, ($\times$) denotes no coverage, and ($\sim$ Partial) denotes limited coverage.

\begin{table}[htbp]
\centering\footnotesize
\caption{Comparative Analysis: Prior Studies vs. Present Study}
\label{tab:comparison}
\resizebox{\textwidth}{!}{%
\begin{tabular}{|p{4.5cm}|c|c|p{3.5cm}|c|c|}
\hline
\textbf{Study / Work} & \textbf{No. of Models} & \textbf{Domain} & \textbf{Metrics} & \textbf{Difficulty Levels} & \textbf{Hallucination Focus}\\
\hline
Brown et al.\ \cite{brown2020gpt3} (GPT-3, 2020)        & 1        & General NLP             & Perplexity, Accuracy                              & $\times$           & $\times$\\
\hline
Liang et al.\ \cite{liang2022helm} (HELM, 2022)          & 30+      & General Multi-task      & 7+ metrics                                        & $\times$           & $\sim$ Partial\\
\hline
Lin et al.\ \cite{lin2022truthfulqa} (TruthfulQA, 2022)  & 10+      & Truthfulness            & Truthful\%, info\%                                & $\times$           & $\checkmark$\\
\hline
Cao et al.\ \cite{cao2023hallucination} (2023)            & Multiple & Survey / LLMs           & Qualitative Taxonomy                              & $\times$           & $\checkmark$\\
\hline
Hendrycks et al.\ \cite{hendrycks2021mmlu} (MMLU, 2021)  & 10+      & Academic Knowledge      & Accuracy                                          & $\sim$ Partial     & $\times$\\
\hline
Maynez et al.\ \cite{maynez2020faithfulness} (2020)      & 3--4     & Text Summarization      & Faithfulness Metrics                              & $\times$           & $\sim$ Partial\\
\hline
Zhang et al.\ \cite{zhang2020bertscore} (BERTScore, 2020) & 5+      & Text Generation         & BERTScore                                         & $\times$           & $\times$\\
\hline
Lewis et al.\ \cite{lewis2020rag} (RAG, 2020)            & Multiple & Knowledge-Intensive NLP & Retrieval Accuracy                                & $\times$           & $\checkmark$\\
\hline
\textbf{Present Study}                                    & \textbf{10} & \textbf{Academic writing} & \textbf{Acc., Prec., Rec., RDI, HI}          & $\checkmark$       & $\checkmark$\\
\hline
\end{tabular}%
}
\end{table}

\noindent\textbf{F. Synthesis and Positioning of the Present Study}

The current literature shows a number of important findings that motivate and
Provide context for the current study\cite{ji2023hallucination}--\cite{nangia2020crows}.
First, most comparative studies consider only a handful of models (usually 2 - 5).
This approach is not practiced at a large scale across different architectures, and the present study seeks to do so, examining ten architectures.
Proprietary and Open-source contemporary LLMs. Second,
Current resources like HELM and MMLU use mostly the same test sets to evaluate.
do not systematically explore the relationship between prompt complexity and model performance.
This study proposes a four-level prompt for and hallucination behavior; accordingly, this study introduces a four level prompt for and hallucination behavior.
A complexity framework of four categories—Basic, Explanation, Advanced, and Tricky—is used.
Third, previous research has been centered on the individual evaluation measures (accuracy), and
the lack of truthfulness, or faithfulness, which leads to an inadequate evaluation of the reliability of the model;
The present work tackles this challenge by suggesting a multimetric evaluation framework
Adding Accuracy, Precision, Recall, Response Deviation Index and Hallucination Index to
comprehensive reliability assessment.

The present study is unique in that it combines the advantages of the two approaches as is shown in the results of the experiments.
The depth of a four-level comparative evaluation is the third dimension of a comparative evaluation.The third dimension of a comparative evaluation is its depth in the case of a four-level comparative evaluation.
prompt complexity framework extends this by incorporating additional metrics from a five-metric set.An extended version of the prompt complexity framework is the comprehensiveness of a five-metric set.
evaluation methodology. To our knowledge this combination has not been
Previously studied thoroughly in the literature. The results are useful for:
The academic use of contemporary LLMs, reliability ranking of these LLMs for academic use and providing.
provided valuable insights into hallucination mitigation, factual alignment, bias reduction and
Improvement of more reliable AI systems.

\section*{III. Methodology}
\label{sec:methodology}

\begin{figure}[htbp]
  \centering
  \includegraphics[width=\textwidth]{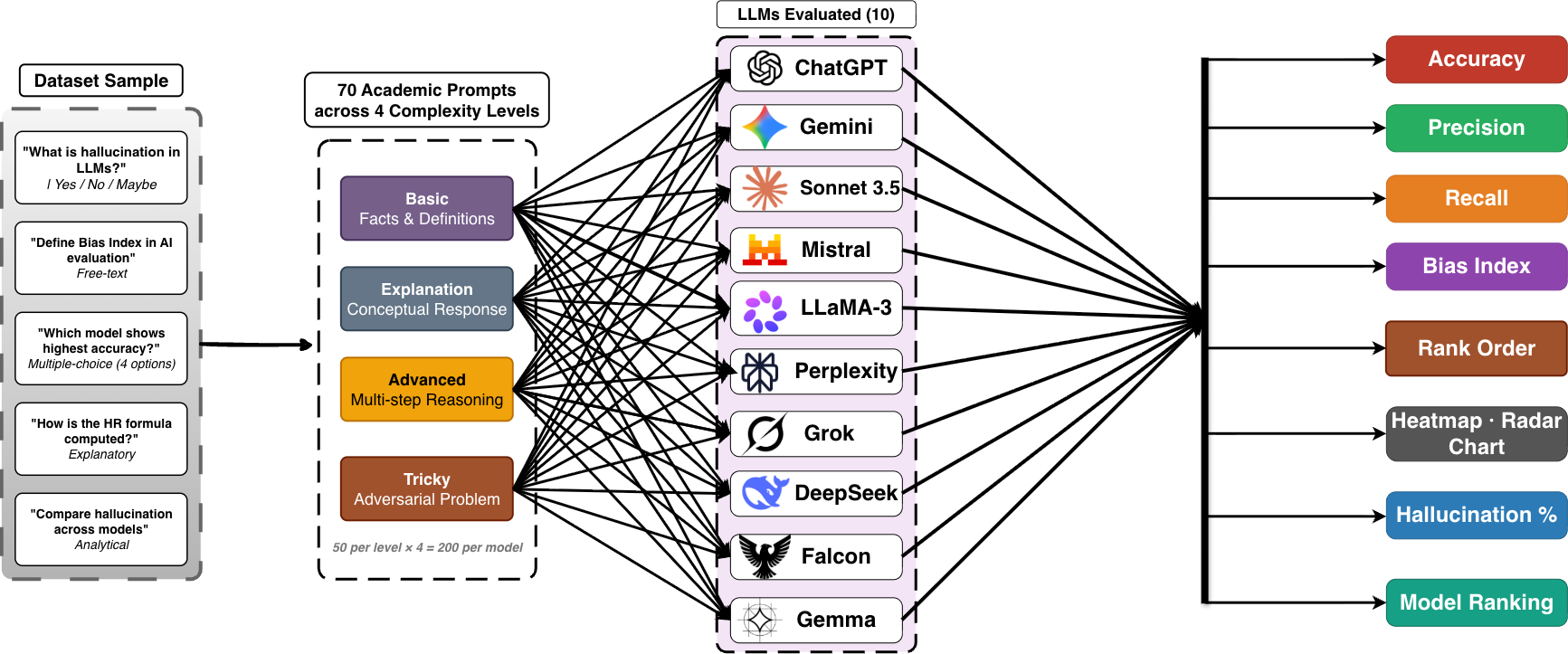}
  \caption{LLM Evaluation Methodology Pipeline}
  \label{fig:methodology}
\end{figure}

\autoref{fig:methodology} Finally presents the full LLM evaluation methodology pipeline,which includes how the benchmark dataset is processed at the four prompt complexity levels,evaluated over ten LLMs under the same conditions,and scored by three independent reviewers using five complementary metrics.

\noindent\textbf{A. Models Evaluated}

This study benchmarked ten cutting-edge Large Language Models (LLMs), covering proprietary and open-sourced models, which are representative of the field's diversity and appeal to both academia and industry. The selected models were identified based on their popularity, design variations, accessibility, and applicability to various domains that LLMs have been applied to. The evaluated LLMs were implemented using different underlying approaches, including the transformer architecture, retrieval-augmented paradigms, or mixture-of-experts designs \cite{zhao2023survey}. All the models were benchmarked under the same set of conditions to guarantee the fairness of comparisons and reproducibility of results by using the default inference parameters for each LLM.

Given the rapid pace of model updates in the field, reproducibility demands that the exact version of each model be documented. All evaluations were conducted in February 2025 using the default inference settings, with no custom system prompts and a single run per prompt per model. \autoref{tab:modelspec} provides the full experimental configuration for each system.

\begin{table}[htbp]
\centering\footnotesize
\caption{Model Specifications and Experimental Configuration}
\label{tab:modelspec}
\resizebox{\textwidth}{!}{%
\begin{tabular}{|l|l|l|c|c|c|c|}
\hline
\textbf{Model} & \textbf{Version / ID} & \textbf{Interface} & \textbf{Access Date} & \textbf{Temp.} & \textbf{Runs} & \textbf{System Prompt}\\
\hline
ChatGPT          & GPT-4o (2024-05)            & chat.openai.com    & Feb 2025 & Default          & 1 & None\\
\hline
Gemini           & Gemini 1.5 Pro              & gemini.google.com  & Feb 2025 & Default          & 1 & None\\
\hline
Claude~3.5 Sonnet & claude-3-5-sonnet-20241022 & claude.ai          & Feb 2025 & Default          & 1 & None\\
\hline
LLaMA~3          & Meta-Llama-3-70B            & Meta AI Web UI     & Feb 2025 & Default          & 1 & None\\
\hline
Mistral~7B       & Mistral-7B-Instruct-v0.3    & chat.mistral.ai    & Feb 2025 & Default          & 1 & None\\
\hline
Grok             & Grok-1.5                    & grok.x.ai          & Feb 2025 & Default          & 1 & None\\
\hline
Perplexity       & Sonar Large Online          & perplexity.ai      & Feb 2025 & Default (RAG ON) & 1 & None\\
\hline
DeepSeek~V3      & deepseek-chat (V3)          & chat.deepseek.com  & Feb 2025 & Default          & 1 & None\\
\hline
Falcon           & Falcon-40B-Instruct         & HuggingFace Spaces & Feb 2025 & Default          & 1 & None\\
\hline
Gemma            & Gemma-7B-IT                 & HuggingFace Spaces & Feb 2025 & Default          & 1 & None\\
\hline
\end{tabular}%
}
\end{table}

\noindent\textbf{1) ChatGPT (OpenAI):}
ChatGPT, which is based on the GPT-4 architecture developed by OpenAI,is one of the most widely adopted proprietary LLMs because of its superior language understanding and generationncapabilities. In the present study, ChatGPT achieved an accuracy of $0.84$,precision of $0.88$, and recall of $0.91$, indicating its strong contextualnunderstanding and response completeness. However, the hallucination index of $12.00\%$ suggests that this model tends to provide factually
unsupported information when responding to complex and adversarial
prompts \cite{bang2023multitask}.

\noindent\textbf{2) Gemini (Google):}
Gemini is Google's foundational model that was constructed for multimodal
reasoning and excels at knowledge-intensive tasks. It has shown the best
performance among all models that have been tested, with a high accuracy rate, which is 0.91, and a great recall rate equal to 0.97. Moreover, it has a relatively low hallucination index, which is 7.84\% \cite{google2023gemini}.Thus, it can be concluded that this model performs the best and is the most consistent one in terms of the different levels of detail in prompts.

\noindent\textbf{3) Claude 3.5 Sonnet (Anthropic):}
Claude~3.5~Sonnet is a proprietary large language model produced by Anthropic that prioritizes safety, alignment, and factual accuracy \cite{anthropic2024claude3}. The model was able to achieve an accuracy of 0.87, precision of 0.90, and a recall of 0.93, achieving one of the lowest
levels of response deviation (RDI = 0.02) of all the evaluated models. Overall, its strong performance profile suggests that it is best suited for academic and professional applications that require a high degree of factual accuracy.

\noindent\textbf{4) LLaMA 3 (Meta AI):}
The LLaMA~3 is an open-source language model created by Meta AI to offer
competitive accuracy while ensuring higher degrees of transparency and
accessibility \cite{meta2024llama3}, \cite{touvron2023llama2}. This model was characterized by an impressive precision score of 0.94 and a low hallucination index of 5.12\%, which implies that it used a conservative generation strategy to produce responses with higher levels of factual accuracy.However, this approach came at the expense of a lower recall value of
0.77.

\noindent\textbf{5) Mistral 7B (Mistral AI):}
Mistral~7B is an efficient open-source language model which achieves a remarkable balance between performance and computational cost through the use of innovative attention mechanisms \cite{jiang2023mistral} and has demonstrated an accuracy of
The precision was 0.84, the recall was 0.91, and the hallucination index was relatively low at 8.69\%. Overall, among all the open-source models examined, Mistral obtained the best balance between factual accuracy and response completeness.

\noindent\textbf{6) Grok (xAI):}
Grok, created by xAI, is intended for use in real-time information retrieval and conversational reasoning \cite{xai2024grok}. While it generated coherent and contextually appropriate responses, it demonstrated the highest hallucination index (12.76\%) of all evaluated systems, suggesting a notable propensity for factual inaccuracies when faced with demanding prompts.

\noindent\textbf{7) Perplexity:}
Perplexity utilizes a retrieval-augmented generation (RAG) paradigm that incorporates external information during inference \cite{perplexity2024}. While the inclusion of retrieval-based information offers benefits,the model achieved a hallucination index of 12.00\%,indicating that merely employing a retrieval-augmented generation framework is insufficient to eliminate hallucinations in sophisticated academic tasks \cite{es2023ragas}.

\noindent\textbf{8) DeepSeek V3:}
DeepSeek~V3 is a large scale Mixture-of-Experts (MoE) language model that offers efficient scaling up and enhanced reasoning \cite{deepseek2024v3}. The model was able to record an accuracy of 0.82 and a recall of 0.89, but it had a very high hallucination index i.e 12.24\% which implies that the model had difficulty maintaining the factual consistency of challenging prompts.

\noindent\textbf{9) Falcon (Technology Innovation Institute):}
Falcon is an open-source language model developed by the Technology Innovation Institute(TII), UAE \cite{tii2023falcon}, \cite{penedo2023refinedweb}. It achieved perfect precision ($1.00$) and a hallucination index of $0.00\%$, revealing superb factual conservatism. However, this conservative strategy led to the lowest recall ($0.64$) and the highest response deviation index ($0.24$), thus making it less suitable for broad academic content production.

\noindent\textbf{10) Gemma (Google DeepMind):}
Gemma is an open-source language model that has been developed through the collaborative work of Google DeepMind and is based on the fundamental architecture of Gemini \cite{google2024gemma}. The model has an accuracy score of 0.74 and a recall rate of 0.70, together with a response deviation index of 0.15, which indicates that it is quite balanced, even if it is not especially strong, and shows no obvious strengths or weaknesses.

\noindent\textbf{B. Prompt Design and Benefits}

To test the hallucination behavior and factual reliability of Large Language Models (LLMs), we have built a benchmark dataset of 70 carefully designed academic prompts. The prompts have been created to cover a wide range of academic writing tasks and have been categorized into four buckets corresponding to an increasing level of complexity aimed at measuring the performance of LLMs when confronted with progressively more difficult tasks \cite{chang2024survey}.

\noindent\textbf{Basic:}This category includes simple factual questions, definitions and simple knowledge retrieval tasks. This category of prompts tests the ability of LLMs to produce correct and well-known information.

\noindent\textbf{Explanation:} This category focuses on conceptual understanding, descriptive reasoning and explanatory answers. The prompts in this category evaluate the LLM's capability to provide coherent and logically structured explanations.
\noindent\textbf{Advanced:} These prompts include analytical reasoning, technical problem solving, and multi-step inference tasks to test the LLM's ability to perform deeper levels of comprehension and contextual reasoning \cite{wei2022cot}.

\noindent\textbf{Tricky:}These are ambiguous and misleading prompts that are intended to trigger hallucination behavior and to highlight possible shortcomings in the factual accuracy and reasoning skills of LLMs \cite{zhu2023dyval}.

The proposed prompt design has several important advantages and benefits. First, it allows to assess factual consistency of LLMs at different levels of task complexity, which offers useful insights into their reliability under increasingly challenging conditions. Secondly, it allows the evaluation of the model robustness against misleading, ambiguous, and adversarial prompts, explores the influence of complexity on the hallucination behavior, and provides recommendations for choosing the most appropriate language model for specific academic writing tasks according to the performance attributes. Finally, the proposed framework provides a standard benchmark methodology,

which can be used as a basis for further research and development efforts to solve and mitigate hallucination-related issues in artificial intelligence.

In order to conduct a thorough evaluation of the performance of the LLMs on the prompts generated, each response has been examined manually and analyzed quantitatively through the use of five different measures of evaluation, including accuracy, precision, recall, Hallucination Index (HI) and Response Deviation Index (RDI), thus providing the scope for a thorough analysis and comparison.

\noindent\textbf{C. Scoring Rubric}

To evaluate Large Language Models (LLMs) for academic content generation, each response was assessed using a standardized scoring rubric that included five evaluation metrics: \textbf{Accuracy (A), Precision (P), Recall (R), Response Deviation Index (RDI), and Hallucination Index (HI)}. Each metric measured
different aspects of performance and reliability of the evaluated models.
Each response was scored on a 10-point scale (0--10), where a score of 0 meant that the response contained severe factual inaccuracies, substantial hallucination, or was of poor quality, while a score of 10 meant that the response had excellent factual correctness, completeness, and consistency. The evaluation focused on five key criteria: factual accuracy, logical consistency, completeness, neutrality, and the absence of fabricated or unsupported information.

To ensure objectivity and minimize subjective bias, three independent evaluators scored each response using a predefined and standardized scoring protocol. Model identities were anonymized to ensure that the reviewers did not know which model they were evaluating. The final score for each metric was calculated as the arithmetic mean of the
individual reviewer scores \cite{zheng2024judging}.

This multi-reviewer evaluation framework increased the reliability, validity, and reproducibility of the comparative analysis and provided a strong basis for evaluating hallucination and factual trustworthiness of contemporary Large
Language Models.

Before presenting results, it is important to clarify how each metric is operationalised. The \textbf{unit of analysis} throughout this study is a single model response to a single prompt, yielding 700 evaluated items in total (10 models $\times$ 70 prompts). Each response was classified relative to its benchmark answer as either a \textbf{True Positive (TP)} a response that correctly addresses the factual content of the benchmark or a \textbf{False Positive (FP)} a response that contains fabricated, inaccurate, or misleading content that cannot be verified against the benchmark. For the Tricky prompts, which embed a false clause alongside a true one, a TP requires the model to correctly identify the false element rather than accepting the compound statement at face value. A \textbf{False Negative (FN)} represents correct benchmark information that the model omitted entirely from its response.

Given this classification, the five metrics are defined as follows. \textbf{Accuracy} measures what proportion of all responses were factually correct:

\begin{equation}
\text{Accuracy} = \frac{TP}{TP + FP + FN}
\label{eq:accuracy}
\end{equation}

\textbf{Precision} captures the reliability of the information a model does produce that is, of everything a model stated, how much of it was actually correct:

\begin{equation}
\text{Precision} = \frac{TP}{TP + FP}
\label{eq:precision}
\end{equation}

\textbf{Recall} measures coverage how much of the correct benchmark content the model successfully retrieved and reported:

\begin{equation}
\text{Recall} = \frac{TP}{TP + FN}
\label{eq:recall}
\end{equation}

The Response Deviation Index (RDI) quantifies the gap between what the model claimed and what was actually true, calculated as the absolute difference between the normalised reviewer score $\hat{y}$ and the ground-truth label $y$. It is explicitly not a measure of social or demographic bias, but a calibration measure of how far model responses deviate from verified reference answers:

\begin{equation}
\text{RDI} = |\hat{y} - y|
\label{eq:rdi}
\end{equation}

The \textbf{Hallucination Index (HI)} specifically isolates the proportion of a model's output that is fabricated or unsupported. Note that it is mathematically equivalent to $(1 - \text{Precision}) \times 100$; it is retained as a separate metric because it communicates the fabrication rate directly as a percentage, without requiring readers to invert the precision value:

\begin{equation}
\text{Hallucination Index (HI)} = \frac{FP}{TP + FP} \times 100
\label{eq:hi}
\end{equation}

To verify that the three reviewers were scoring consistently rather than arbitrarily, the \textbf{Intraclass Correlation Coefficient} was computed for each metric using a two-way mixed-effects model with absolute agreement, ICC(3,1). The resulting values were: Accuracy = 0.81, Precision = 0.79, Recall = 0.83, Response Deviation Index = 0.76, and Hallucination Index = 0.84. All figures comfortably exceed the 0.75 threshold associated with good reliability \cite{koo2016guideline}. \textbf{Krippendorff's $\alpha$} was also computed for the ordinal rating scale and returned $\alpha = 0.79$, corroborating that the level of agreement among reviewers was substantial and not attributable to chance.

\noindent\textbf{D. Hallucination Index}

In order to quantitatively evaluate hallucination behavior, in this paper, we define a new metric, called \textbf{Hallucination Index (HI)}, which measures the fraction of factually incorrect, unsupported, or fabricated information produced by a Large Language Model (LLM). A response is considered hallucinated if it contains inaccurate facts,references to non-existent entities, misleading statements, or unsubstantiated claims that cannot be corroborated by reliable sources \cite{manakul2023selfcheckgpt}, \cite{min2023factscore}.

\noindent\textbf{The Hallucination Index is computed using the following formulation:}

\noindent Hallucination Rate (HR) Formula:

\begin{center}
$HR = \dfrac{FP}{(TP+FP)} \times 100$
\end{center}

\noindent\textbf{Where,}

\noindent\hspace{2cm}\textbf{HR} = Hallucination rate (In Percentage)

\noindent\hspace{2cm}\textbf{FP} = False positive (Incorrect or hallucinated Information)

\noindent\hspace{2cm}\textbf{TP} = True positive (Correct and factual Information)

\begin{figure}[htbp]
  \centering
  \includegraphics[width=0.90\textwidth]{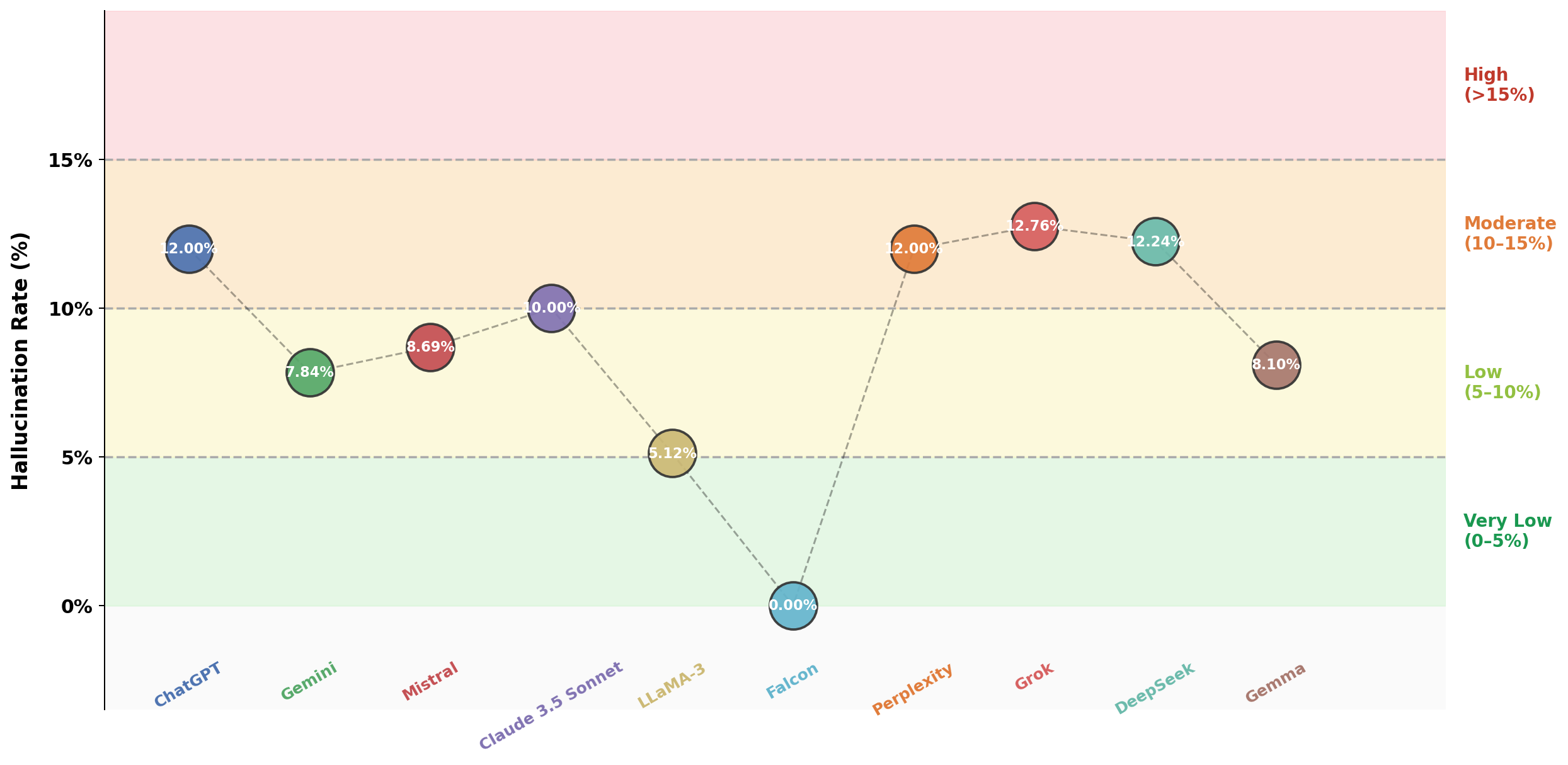}
  \caption{Model Positions Hallucination Index (HI) Score Classification Scale}
  \label{fig:hiscale}
\end{figure}

The Hallucination Index therefore shows what proportion of the generated contents are hallucinated among all the factual statements generated; the lower the index the more factual and reliable the results are, while a higher one means the contents are more prone to hallucination. For reference, the grading scale is given below in \autoref{fig:hiscale},
which positions all ten evaluated models against four hallucination zones Very Low,
According to their measured hallucination rates, the levels are classified as Low, Moderate, and High.

The Hallucination Index is particularly valuable for evaluating academic writing tasks,
Since factual accuracy, the validity of citations, and the trustworthiness of the content are important \cite{alkaissi2023artificial}, \cite{athaluri2023exploring}. When
combined with \textbf{Accuracy}, \textbf{Precision}, \textbf{Recall}, and \textbf{Response Deviation Index}, the Hallucination Index provides a comprehensive framework for assessing the
reliability, factual consistency, and practical applicability of contemporary Large
Language Models.

\begin{figure}[htbp]
  \centering
  \includegraphics[width=\textwidth]{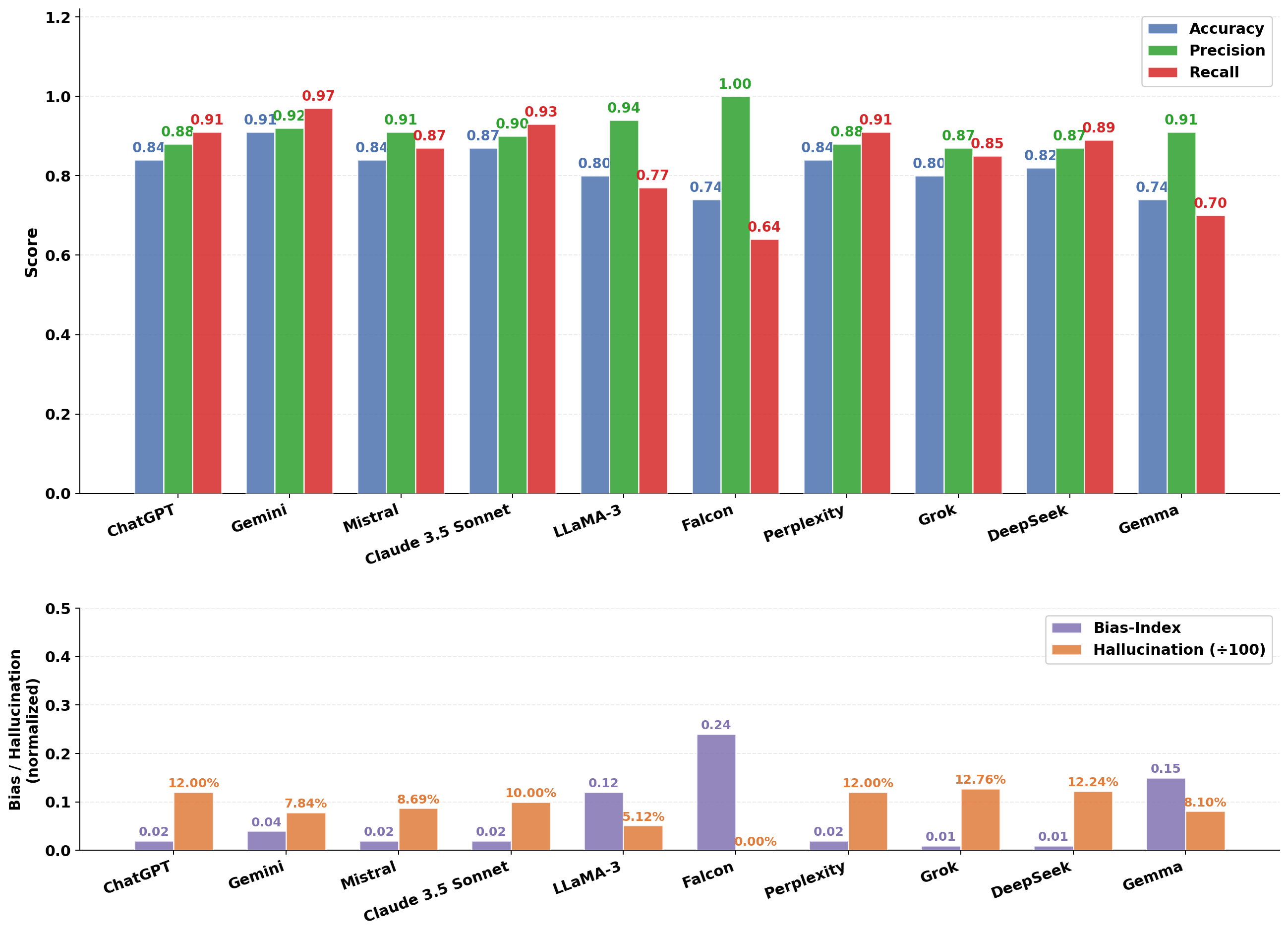}
  \caption{Comparative Performance Analysis of Large Language Models}
  \label{fig:performance}
\end{figure}

\autoref{fig:performance} presents the comparative performance analysis of all ten LLMs
across all five evaluation metrics simultaneously, including Accuracy, Precision, Recall,
Response Deviation Index, and Hallucination rate.

\section*{IV. Experimental Results}
\label{sec:results}

\noindent\textbf{A. Overall Performance of the Models}

\autoref{fig:performance} presents the comparison of performance of ten evaluated Large Language Models (LLMs) based on five assessment criteria, namely Accuracy,Precision, Recall, Response Deviation Index and Hallucination Index. It can be seen that all LLMs demonstrate quite different abilities in terms of factual accuracy of responses,completeness of answers, presence of biases and hallucination.

Among all the evaluated models, \textbf{Gemini} demonstrated the best performance achieving the highest accuracy (0.91) and recall (0.97) and a relatively low hallucination index (7.84\%). Thus, Gemini has the best overall performance among all the LLMs tested in this study.

\textbf{Claude~3.5~Sonnet} demonstrated the most balanced performance with accuracy,precision, recall, and a low response deviation index of 0.87, 0.9, 0.93 and 0.02 respectively.As for the open-source models, \textbf{Mistral} achieved the best overall performance with accuracy, precision, recall, and hallucination index of 0.84, 0.91, 0.87 and 8.69\%.

\textbf{LLaMA~3} demonstrated a high precision (0.94) and a relatively low hallucination index (5.12\%), which suggests that it provides the most accurate answers with the lowest possibility of fabricating information.
\textbf{ChatGPT} and \textbf{Perplexity} demonstrated quite similar performance achieving accuracy and recall of 0.84 and 0.91, respectively, with a relatively high hallucination index of 12\%. Both models demonstrated high ability to generate coherent and contextually consistent responses, however, they were not accurate enough and prone to hallucination when answering challenging questions \cite{qin2023chatgpt}.

\textbf{Grok} and \textbf{DeepSeek~V3} demonstrated similar patterns, achieving almost the same hallucination index of 12.76\% and 12.24\%, respectively. Both models were quite good at providing consistent and coherent answers, however, their responses were not accurate enough, especially when answering challenging questions.

\textbf{Falcon} demonstrated a different pattern achieving perfect precision of 1 and the lowest hallucination index of 0\%. However, the model failed to achieve high recall (0.64) and demonstrated a moderately high response deviation index of 0.24. Thus, although Falcon was not prone to hallucination and could provide factually accurate answers,it avoided any responses that could be considered potentially biased.Finally, \textbf{Gemma} demonstrated the lowest performance achieving accuracy and recall of 0.74 and 0.7, respectively. Gemma demonstrated a moderate response deviation index (0.15) and was outperformed by all other models in all the metrics.

Based on the study, it can be concluded that no model demonstrated the best
performance in all the metrics. However, \textbf{Gemini}
and\textbf{Claude~3.5~Sonnet} demonstrated the best overall performance achieving a relatively high accuracy and recall and low levels of hallucination and response deviation. Other models demonstrated more mixed performance with strengths and weaknesses in different areas. In addition,although some models demonstrated high accuracy-recall performance, they generally
had a higher hallucination index which indicates that they might not be suitable for providing factual information in research or academic writing. The results also suggest that the tendency to hallucinate is not necessarily related to the general performance of the models. Rather, different models have different strengths and weaknesses in terms of their ability to avoid response deviation and hallucination.

\noindent\textbf{B. Performance Across Five Evaluation Metrics}

\autoref{fig:performance} presents a comprehensive comparison of performances of the ten Large Language Models (LLMs) on the five criteria: Accuracy, Precision, Recall, Response Deviation Index, and Hallucination Index. The results suggest the trade-offs between factual correctness, response completeness, response deviation, and hallucination.

In terms of the Accuracy and Recall, Gemini
achieved the best results with $0.91$ and $0.97$,
respectively, while Claude~3\.5~Sonnet recorded $0.87$ in
accuracy and $0.93$ in recall, which can be considered as a moderate
result. Mistral, an open-source model, is competitive with
$0.84$ in accuracy and $0.87$ in recall.

In terms of Precision, Falcon attained the best result of $1.00$, followed by LLaMA~3 with $0.94$. However,both Falcon and LLaMA~3 recorded relatively low Recall ($0.64$, $0.77$, respectively), which implies their conservativeness in response generation.

The Response Deviation Index indicates that some models were more neutral and
consistent in their responses than others. Indeed, Grok,
DeepSeek~V3 recorded the lowest response deviation ($0.01$), while
Falcon reported the highest ($0.24$), indicating the
impact of using the most restrictive generation policy \cite{gallegos2024bias}.

The Hallucination Index suggests that the extent to which the
models generate information not present in the input varies considerably.
Falcon was the least likely to have hallucination
($0.00\%$), followed by LLaMA~3 ($5.12\%$) and
Gemini ($7.84\%$). Grok ($12.76\%$),
DeepSeek~V3 ($12.24\%$), ChatGPT
($12.00\%$), Perplexity ($12.00\%$) were more
prone to having hallucination, which might impact their performance on
complex academic writing tasks \cite{rawte2023survey}.

\begin{figure}[htbp]
  \centering
  \includegraphics[width=\textwidth]{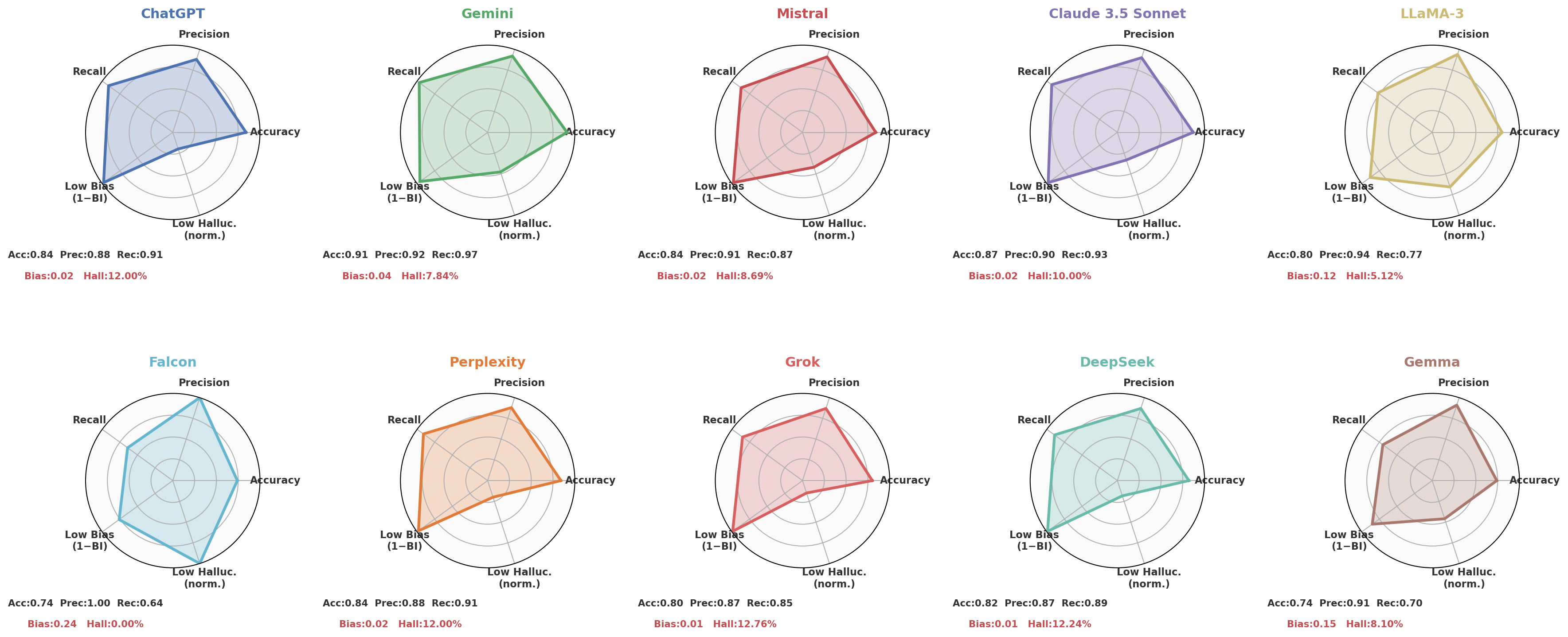}
  \caption{Detailed Individual Model Performance Profiles}
  \label{fig:detailed}
\end{figure}

One can also observe a trade-off between Recall and
Hallucination Index. One can see that higher Recall is
associated with increased chances of hallucination. At the same time,
higher Precision does not necessarily guarantee high overall performance,
as seen in Falcon's case. Thus, no single metric is sufficient
for evaluating LLMs' performance on the given academic writing task.
The results for all the models on the five criteria are presented in \autoref{fig:detailed}.

\FloatBarrier

\noindent\textbf{C. Hallucination Rate Analysis Across Models}

\begin{figure}[htbp]
  \centering
  \includegraphics[width=\textwidth]{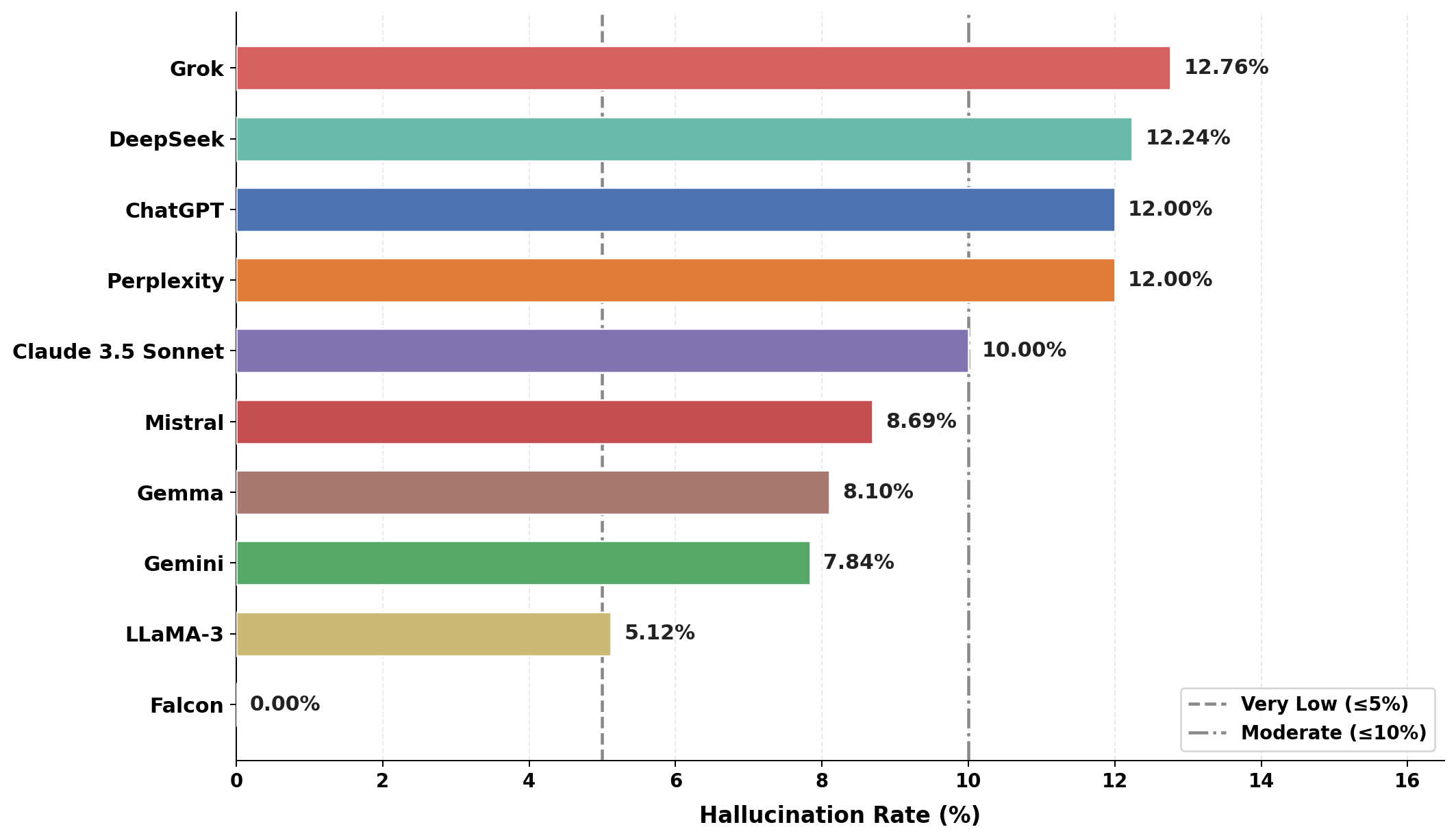}
  \caption{Overall Hallucination Rate (HR) per Model}
  \label{fig:hrbar}
\end{figure}

The dissimilarity in factuality preservation during the academic writing tasks is illustrated in the comparative Hallucination Index (HI) for the evaluated Large Language Models in \autoref{fig:hrbar}. The results suggest that hallucination is mainly influenced by the underlying architecture, alignment strategies and response generation mechanisms \cite{huang2023survey_hallucination}.

\textbf{Falcon} had the lowest hallucination rate (0.00\%), followed by LLaMA~3 (5.12\%), indicating strong resistance to producing fabricated or unsupported information. However, this improved hallucination control comes at the cost of lower recall values (0.64 and 0.77 respectively), indicating that these models employ conservative generation strategies focusing on factual safety over response completeness.

And other models had relatively low hallucination rate, yet high precision and recall. It indicates a good perception of factuality and generation of response that would make the models fit well for educational and knowledge-intensive applications. The ChatGPT (12.00\%), Perplexity (12.00\%),DeepSeek~V3 (12.24\%), and Grok (12.76\%) ranked on the highest in terms of hallucination rate among the tested models. 
They all produce very convincing, rich in information, coherent, and natural responses, but with an increased appearance of plausible but ungrounded statements, especially, when dealing with sophisticated, confusing, or adversarial input. 
The main takeaway is a construction of a trade-off between \textbf{hallucination resistance} and \textbf{response granularity}, whereas models emphasizing comprehensive and detailed outputs exhibited higher hallucination tendencies \cite{peng2023check}.

\FloatBarrier

\noindent\textbf{D. Frequency of Low-Quality Responses}

The frequency of low-quality responses serves as an essential indicator of the model ro-

bustness, consistency, and factual reliability under different degrees of prompt complexity.Analysis of the results presented in \autoref{fig:hrbar} reveals substantial differences in the ability of Large Language Models (LLMs) to maintain response quality across academic writing tasks.

Gemini, Claude~3.5~Sonnet, and Mistral achieved the highest levels of stability by

producing the highest quality responses with the highest precision-recall values,
and the lowest hallucination rates, suggesting that they are the most factually sound models among the evaluated ones, which produced the lowest amount of low-quality responses.
LLaMA~3 and Falcon achieved moderate levels of stability, with relatively low
hallucination rates and high precision, but low recall due to the models' tendency
to refrain from producing responses that may contain facts that are not supported
by the context.

Grok, DeepSeek~V3, ChatGPT, and Perplexity had the highest rates of low-quality
responses, with hallucination rates ranging from 12.00\% to 12.76\%.
Overall, it can be stated that the higher the hallucination rate, the lower the quality of responses produced by a model. In addition, the analysis revealed that
as the complexity of the prompts increased, the quality of responses decreased,resulting in more low-quality responses. The highest quality of responses was observed for Basic and Explanation prompts, with the quality dropping considerably for Advanced and Tricky prompts.An interesting observation of the current analysis is that the frequency of hallucinations appeared to have a considerable impact on the quality of responses, with higher indices being associated with lower quality, supporting the initial assumptions about the impact of hallucinations on response quality. Thus, it can be concluded that the most stable models were the ones that did not sacrifice the quality of responses for the increased level of detail and precision, which is critical for responses in languages other than English.

\noindent\textbf{E. Effect of Prompt Complexity on Hallucination: Statistical Analysis}

The preceding sections describe performance patterns across models, but the question of whether prompt complexity \textit{significantly} drives those differences calls for a formal test rather than descriptive comparison alone. A Kruskal--Wallis $H$-test was therefore applied to the per-question error rates (computed as $1 -$ accuracy) across the four complexity levels. This non-parametric approach was selected because the per-question error distributions are ordinal and depart from normality. The test returned:

\begin{equation}
H(3) = 35.21,\quad p < 0.001,\quad \eta^2 = 0.50
\end{equation}

The result is clear: error rates vary significantly across the four levels, and the effect size is large ($\eta^2 = 0.50$), meaning that prompt complexity alone accounts for half the observed variance in error rates. \autoref{tab:complexity} breaks down mean accuracy by model and complexity level, making the gradient visible across all ten systems.

\begin{table}[htbp]
\centering\footnotesize\setlength{\tabcolsep}{4pt}
\caption{Mean Accuracy by Model and Prompt Complexity Level}
\label{tab:complexity}
\resizebox{\textwidth}{!}{%
\begin{tabular}{|l|c|c|c|c|c|}
\hline
\textbf{Model} & \textbf{Basic (n=20)} & \textbf{Explanation (n=17)} & \textbf{Advanced (n=13)} & \textbf{Tricky (n=20)} & \textbf{Overall}\\
\hline
Gemini            & 0.950 & 1.000 & 0.538 & 1.000 & 0.900\\
\hline
ChatGPT           & 0.850 & 0.941 & 0.846 & 0.950 & 0.900\\
\hline
Mistral           & 0.850 & 0.941 & 0.692 & 1.000 & 0.886\\
\hline
Perplexity        & 0.950 & 0.941 & 0.846 & 0.950 & 0.929\\
\hline
Grok              & 0.950 & 0.882 & 0.846 & 0.800 & 0.871\\
\hline
Claude~3.5 Sonnet & 0.700 & 1.000 & 0.538 & 1.000 & 0.829\\
\hline
LLaMA~3           & 0.600 & 0.882 & 0.462 & 0.950 & 0.743\\
\hline
Falcon            & 0.650 & 0.176 & 0.385 & 0.950 & 0.571\\
\hline
DeepSeek~V3       & 0.900 & 1.000 & 0.846 & 0.950 & 0.929\\
\hline
Gemma             & 0.700 & 0.412 & 0.077 & 1.000 & 0.600\\
\hline
\textbf{Mean}     & 0.810 & 0.818 & 0.608 & 0.955 & 0.798\\
\hline
\end{tabular}%
}
\end{table}

Post-hoc pairwise Mann--Whitney $U$ tests with Bonferroni correction showed that the Tricky level stood apart from all other three levels: Basic vs.\ Tricky ($U = 310.5$, $p < 0.001$), Explanation vs.\ Tricky ($U = 301.5$, $p < 0.001$), and Advanced vs.\ Tricky ($U = 274.5$, $p < 0.001$). The three non-Tricky levels, by contrast, did not differ significantly from one another ($p > 0.39$). What this tells us is that Basic, Explanation, and Advanced prompts challenge models in broadly similar ways, but the Tricky prompts which embed a factually false clause within an otherwise plausible statement demand a qualitatively different kind of reasoning. Models that rely on surface-level pattern matching tend to accept the compound statement without scrutinising the embedded falsehood, and this is where hallucination rates climb most sharply.

\section*{V. Discussion}
\label{sec:discussion}
Looking at the ten evaluated Large Language Models shows.They differ a great deal in their ability to maintain factual accuracy and in the quality of their responses. Completeness, bias-free output, and resistance to hallucinations — all measured through a multi The model used in this study shows that performance alone isn't enough to fully All the models have their own particular advantages, which is why they are assessed on the basis of a single metric and limitations Of the models that were evaluated, \textbf{Gemini} and \textbf{Claude 3.5 Sonnet} reached the highest level of accuracy.The performance is mostly well-balanced, with consistently good accuracy and precision and make sure to remember the information while keeping the number of hallucinations as low as possible. They have shown a strong performance points to how powerful advanced alignment methods and solid factual grounding can be They are particularly well suited to producing academic content.

\textbf{Mistral} and \textbf{LLaMA 3} also showed strong performance,
especially when it comes to precision and avoiding false outputs—particularly LLaMA~3
It achieved one of the lowest hallucination rates without losing precision.
favoring a cautious approach that sticks to verified facts
It is more important to get the facts correct than to have every detail right. In a similar way, \textbf{Falcon} demonstrated exceptional
resistant to hallucinations and extremely precise, though it struggled with much lower performance,It has a tendency to forget things and places too much reliance on assumptions, which makes it unsuitable for use in serious academic work.

In contrast, \textbf{ChatGPT}, \textbf{Grok}, and \textbf{DeepSeek}
Displayed a great deal more hallucinations. These results indicate that models Since it is designed for generating detailed and rich text, it might become more susceptible to producing inaccurate or misleading content under adversarial prompting conditions \cite{bang2023multitask}, \cite{shahriar2024putting}.

A key result of this study is that there is a compromise between the completeness of the models' responses and their resistance to hallucination models which produce responses that are highly informative and detailed are found to carry a much greater risk of hallucinating, whereas those that are good at resisting hallucinations tend to give conservative and frequently incomplete responses. Furthermore, our experiments indicate that the complexity of the prompt has a significant impact on the model's reliability – greater complexity in the prompting leads to a marked rise in the frequency of hallucinations \cite{rawte2023survey}, \cite{huang2023survey_hallucination}. In general, our results show that the factual reliability of Large Language Models is governed by a combination of model characteristics, including alignment strategies, response generation mechanisms, and general robustness.The proposed multi-metric evaluation framework is essential for benchmarking LLMs' factual reliability and gives important guidance for the development of more reliable and factually accurate AI assistants \cite{sun2024trustllm}.

\begin{figure}[htbp]
  \centering
  \includegraphics[width=0.85\textwidth]{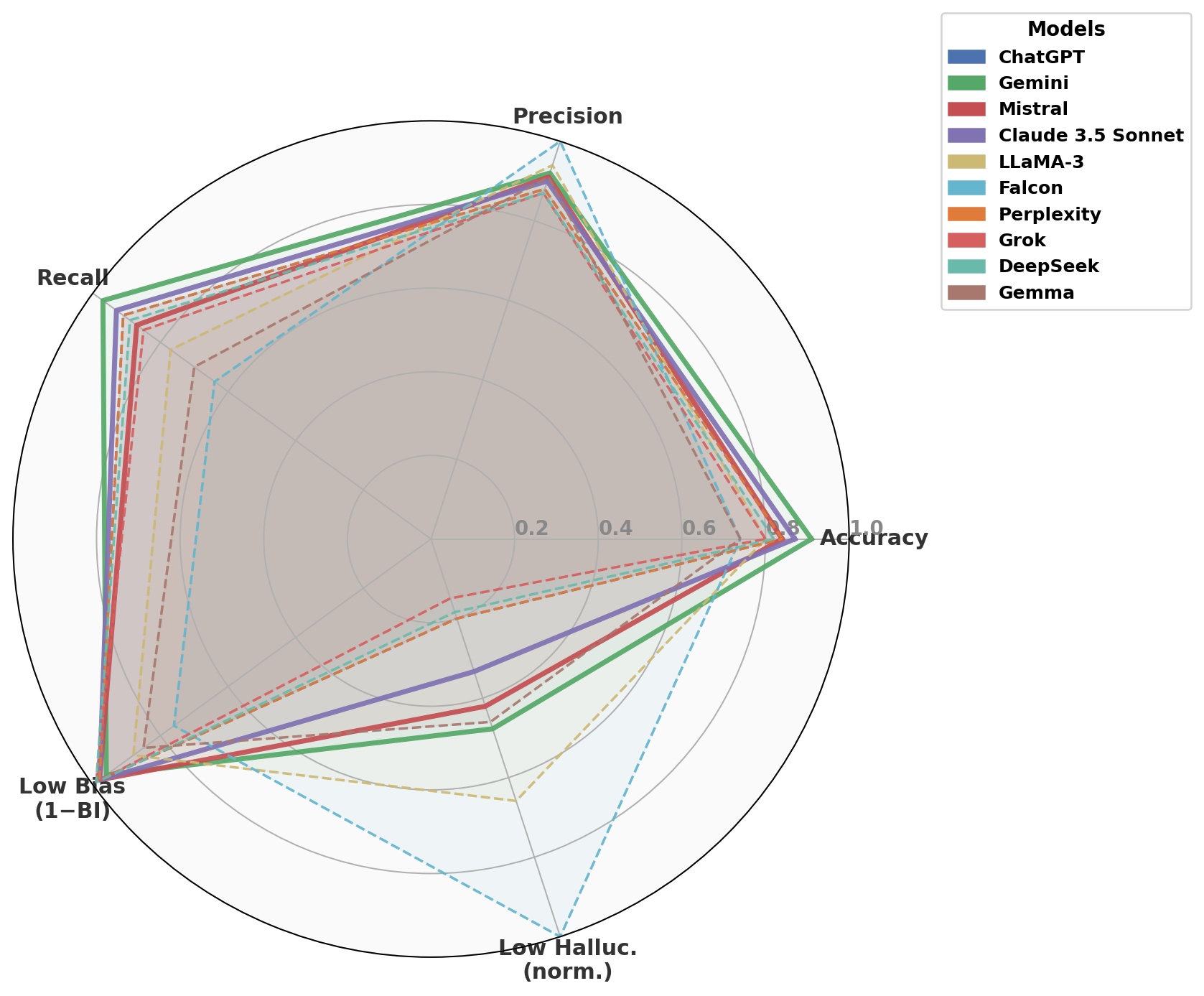}
  \caption{Performance Profile Radar -- 10 LLMs across 5 Metrics}
  \label{fig:radar}
\end{figure}

\autoref{fig:radar} shows the radar chart of performance profiles comparing all ten models simultaneously across the five evaluation dimensions, providing a holistic view of the strengths and weaknesses of each model.

\begin{figure}[htbp]
  \centering
  \includegraphics[width=\textwidth]{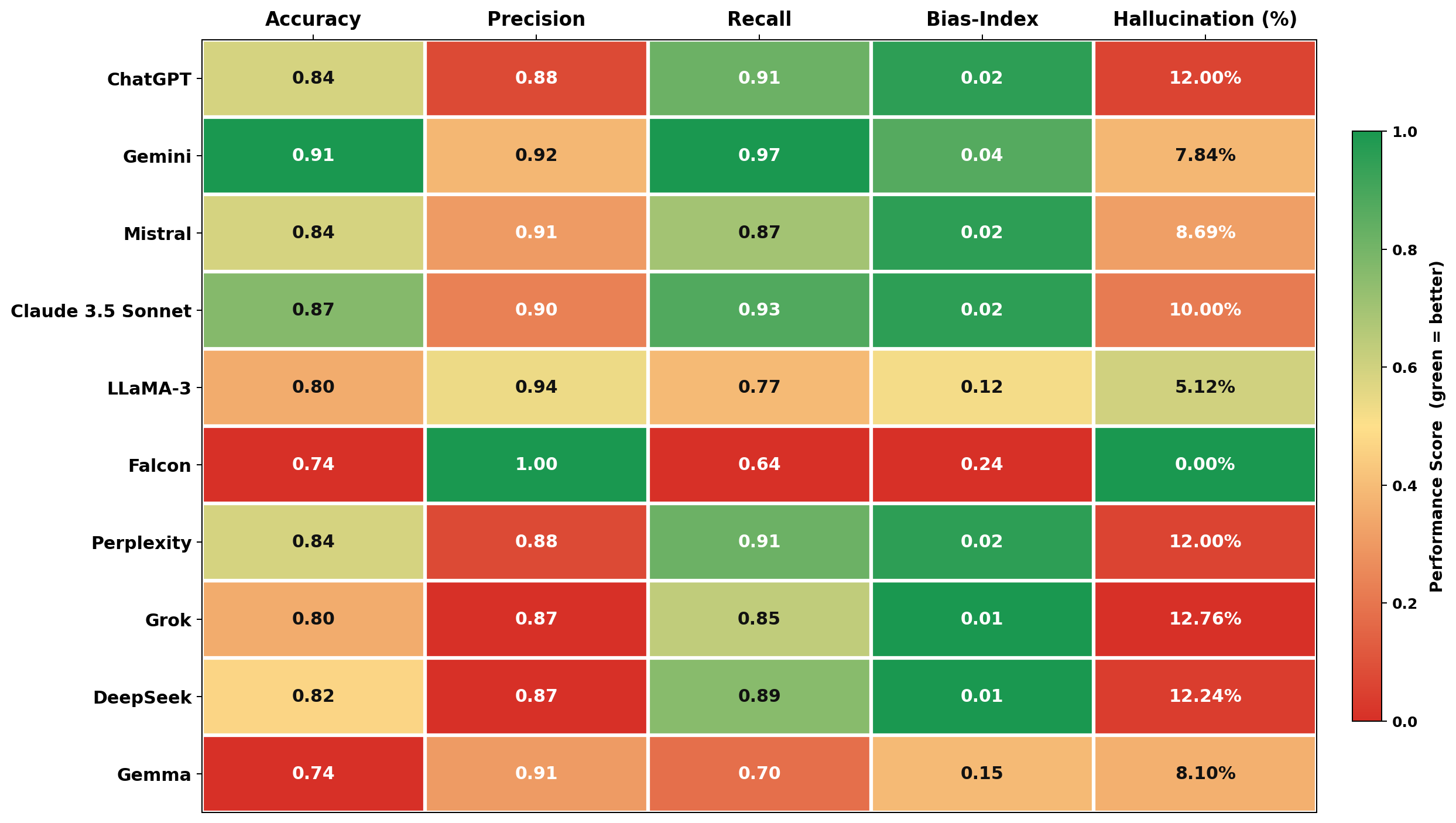}
  \caption{LLM Performance Heatmap - Metric Breakdown in Detail}
  \label{fig:heatmap}
\end{figure}

\autoref{fig:heatmap} Visually presents a complete comparison of 10 Large Language Models (LLMs) with respect to 5 metrics: Accuracy, Precision, Recall, Response Deviation Index and Hallucination Rate through heatmaps. With colour coding it will be easy to see how well the model is working (green = better performance and less risk; red = worse performance, and more risk of bias or hallucination). The heatmap shows that no single model performs well on all evaluation metrics, indicating that there are inherent trade-offs in existing models of LLM architecture. 

Overall, Gemini achieves the best performance among all the models, with the highest Accuracy ( 0.91) and Recall ( 0.97) and relatively low Hallucination Rate ( 7.84\%). In a similar fashion, Claude 3.5 Sonnet exhibits a well-rounded performance profile with high factual accuracy, high recall, low response deviation, and low hallucination, making it one of the most reliable models for academic purposes. Mistral is the best balance between performance and reliability of the open source models, while LLaMA~3 provides high precision with minimal hallucinations through a more cautious generation approach. 

By comparison, ChatGPT, Perplexity, Grok, and DeepSeek V3 exhibit relatively high hallucination rates, indicating that effective language skills and logical reasoning are not sufficient to guarantee factual accuracy. Gemma has moderate performance across all scores, while Falcon has the highest scores in precision and lowest in hallucation, but lower scores in recall and high response deviation, meaning it is not as useful for general academic purposes. 

The multi-metric assessment framework's summary of the overall ranking of the models is Gemini, Claude~3.5~Sonnet, Mistral, Perplexity, ChatGPT, DeepSeek~V3, Grok, LLaMA~3, Gemma, and Falcon. 

Overall, the results of the heatmap analysis give rise to three important observations: (1) linguistic fluency does not always correlate with factual accuracy; ( 2) a lack of facts and the control of hallucination are not mutually independent; ( 3) a single metric is not enough to gauge model reliability. The results highlight the importance of employing multi-metric evaluation tools when evaluating LLMs in academic and other significant settings \cite{chang2024survey}, \cite{liu2023geval}.

\FloatBarrier

\section*{VI. Conclusion}
\label{sec:conclusion}

In this work, we present a thorough multi-metric comparative analysis of ten state-of-the-art Large Language Models (LLMs) for academic content generation, using a benchmark of 70 prompts of four levels of difficulty: Basic, Explanation, Advanced, and Tricky. All evaluations were conducted in February 2025 on specific model versions; findings should be interpreted as a comparative snapshot of those systems rather than a definitive ranking of current capabilities. Experimental results showed large differences in the performance of the models, their factuality, response deviation, and hallucination behavior indicating that no model is best in all evaluation dimensions.

Of the models we looked at, \textbf{Gemini} and \textbf{Claude~3.5~Sonnet} were the most robust and generally capable systems, consistently achieving high accuracy, high recall, low response deviation, and reasonable hallucination rates. Mistral was the strongest open-source option, with the good tradeoff between factual correctness and completeness of the response. In contrast, LLaMA~3 and Falcon were surprisingly resistant to hallucination due to conservative response generation strategies, but their low recall metrics restricted their use in comprehensive academic tasks. Despite the fluency and coherence of outputs, models such as ChatGPT, Perplexity, DeepSeek~V3 and Grok , demonstrated relatively higher rates of hallucination confirming that linguistic fluency is not a guaranty of factual accuracy.

We also find that as prompt complexity increases, the frequency of hallucinations increases and response quality decreases for all models that we evaluate. In particular, the Advanced and Tricky prompts showed significant shortcomings in reasoning, factual grounding, and reliability, highlighting the role of task complexity on the LLMs' performance. It is also important to note that these findings reflect specific model versions evaluated in February 2025; newer model versions may exhibit different performance characteristics. Furthermore, each model was queried once per prompt, which limits claims about generation robustness across multiple runs.

In brief, this work reaffirms that hallucination remains a systemic challenge in modern LLMs, and points to fundamental trade-offs between factual accuracy, response completeness and hallucination control \cite{ji2023hallucination}, \cite{rawte2023survey}. Our proposed five-metric evaluation framework of Accuracy, Precision, Recall, Response Deviation Index, and Hallucination Index offers an extensive and practical approach to evaluate the trustworthiness of LLMs in academic and high-stakes applications \cite{sun2024trustllm}, \cite{wang2023survey_trustworthy}.

Future work should focus on better factual grounding techniques, retrieval-augmented and hybrid architectures, real-time knowledge verification mechanisms, and advanced hallucination mitigation techniques to improve the reliability, transparency and trustworthiness of next-generation Large Language Models \cite{peng2023check}, \cite{shuster2021retrieval}.
\section*{Acknowledgement}

The authors declare that this research has been undertaken in adherence with
Developed rules of academic honesty, research ethics and publication ethics.
In this study, the use of Large Language Models (LLMs) was limited to
restricted to supportive tasks such as language refinement and grammatical corrections,
Formatting, and editing help. At no point were AI tools used to:
Replace the original scientific thinking, experimental design, data analysis, interpretation,
or intellectual contribution.

All benchmark data sets, experimental evaluations, comparative analyses, visualisations,
The design, verification and conclusions presented in this work are independent of those of the other authors.
Approve by the authors. The authors assume full responsibility for the accuracy,
The authenticity, repeatability and integrity of the research results presented in this
study.

The authors would also like to thank the open source and proprietary contributions,
Large Language Model communities, whose developments have made it possible to compare them in meaningful ways.
hallucination analysis, bias evaluation, and evaluation of generalisation ability.
trustworthy artificial intelligence systems. This work aims to contribute toward the
The advancement of trustworthy, transparent, and factually based AI technologies.
academic and professional applications.
\bibliographystyle{IEEEtran}
\bibliography{references}

\end{document}